\documentclass[11pt,a4paper]{article} 

\usepackage[a4paper,left=2cm, right=2cm, top=2.5cm, bottom=2.5cm]{geometry}
\usepackage[T1]{fontenc}
\usepackage{graphicx}
\usepackage{xcolor}
\usepackage{booktabs}
\definecolor{myblue}{RGB}{25,25,112}

\usepackage[english]{babel} 
\usepackage{natbib}
\usepackage[
    colorlinks,
    linkcolor=myblue!80!blue,
    citecolor=myblue!80!blue,
    urlcolor=myblue!60!blue
    ]{hyperref}
 
\usepackage{authblk}
\usepackage{amsmath,amssymb, amsthm}

\usepackage{caption}
\usepackage{url}

\usepackage{bm}

\title{Large-scale benchmark study of survival prediction methods using multi-omics data}

\title{Large-scale benchmark study of survival prediction methods using multi-omics data}

\author[1]{Moritz Herrmann}
\author[2]{Philipp Probst}
\author[2]{Roman Hornung}
\author[2]{Vindi Jurinovic}
\author[2]{Anne-Laure Boulesteix}
\affil[1]{Department of Statistics, LMU Munich, Munich, Germany}
\affil[2]{Institute for Medical Information Processing, Biometry, and Epidemiology, LMU Munich, Munich, Germany}

\date{}

\begin{document}

\maketitle

\abstract{Multi-omics data, that is, datasets containing different types of high-dimensional molecular variables (often in addition to classical clinical variables), are increasingly generated for the investigation of various diseases. Nevertheless, questions remain regarding the usefulness of multi-omics data for the prediction of disease outcomes such as survival time. It is also unclear which methods are most appropriate to derive such prediction models. We aim to give some answers to these questions by means of a large-scale benchmark study using real data. Different prediction methods from machine learning and statistics were applied on 18 multi-omics cancer datasets from the database "The Cancer Genome Atlas", containing from 35 to 1,000 observations and from 60,000 to 100,000 variables. The considered outcome was the (censored) survival time. Twelve methods based on boosting, penalized regression and random forest were compared, comprising both methods that do and that do not take the group structure of the omics variables into account. The Kaplan-Meier estimate and a Cox model using only clinical variables were used as reference methods. The methods were compared using several repetitions of 5-fold cross-validation. Uno's C-index and the integrated Brier-score served as performance metrics. The results show that, although multi-omics data can improve the prediction performance, this is not generally the case. Only the method block forest slightly outperformed the Cox model on average over all datasets. Taking into account the multi-omics structure improves the predictive performance and protects variables in low-dimensional groups---especially clinical variables---from not being included in the model. All analyses are reproducible using freely available R code.}

\textit{Keywords:} multi-omics data, prediction models, benchmark, survival analysis, machine learning, statistics

\section{Introduction}

In the last two decades, high-throughput technologies have made data stemming from molecular processes available on a large scale (\lq\lq omics data'') and for many patients. Starting from the analysis of whole genomes, other molecular entities such as mRNA or peptides have also come into focus with the advancing technologies. 
Thus, various types of omics variables are currently under investigation across several disciplines such as genomics, epigenomics, transcriptomics, proteomics, metabolomics, and microbiomics \citep{Hasin2017}.

It may be beneficial to include these different data types in models predicting outcomes, such as the survival time of patients.
Until recently, only data from a single omics type was used to build such prediction models,  with or without inclusion of standard clinical data \citep{Boulesteix2011}. 
In recent years, however, the increasing availability of different types of omics data measured for the same patients (called \lq\lq multi-omics data'' from now on) has led to their combined use for building outcome prediction models. 
An important characteristic of multi-omics data is the high-dimensionality of the datasets, which frequently have more than 10,000 or even 100,000 variables. This places particular demands on the methods used to build prediction models: they must be able to handle data where the number of variables by far exceeds the number of observations. Moreover, practitioners often prefer sparse and interpretable models containing only a few variables \citep{Klau2018}. Last but not least, multi-omics data are structured: the variables are partitioned into (non-overlapping) groups. This structure may be taken into account when building prediction models. 

Several methods have been specifically proposed to handle multi-omics data, while established methods for high-dimensional data from the fields of statistics and machine learning also seem reasonable for use in this context.
Although there are studies with a limited scope comparing some of these methods, there has not yet been a large-scale systematic comparison of their pros and cons in the context of multi-omics using a sufficiently large amount of real data.

The pioneering study by \citet*{Bovelstad2009} investigates the combined use of clinical and one type of molecular data, using only four datasets. In one of the first studies devoted to methodological aspects of multi-omics-based prediction models, \citet{Zhao2014} compare a limited number of methods for multi-omics data based on a limited number of datasets. 
\citet{LangKotthaus2015} investigate automatic model selection in high-dimensional survival settings, using similar but fewer prediction methods than our study. Moreover, again only four datasets are used. A study by \citet{Bin2019} investigates the combination of clinical and molecular data, with a focus on the influence of correlation structures of the feature groups, but it is based on simulated data.

Our study aims to fill this gap by providing a large-scale benchmark experiment for prediction methods using multi-omics data. It is based on 18 cancer datasets from the Cancer Genome Atlas (TCGA) and focuses on survival time prediction. We use several variants of three widely used modeling approaches from the fields of statistics and machine learning: penalized regression, statistical boosting, and random forest. The aim is to assess the performances of the methods, the different ways to take the multi-omics structure into account, and  the added predictive value of multi-omics data over models using only clinical variables. 

The remainder of the paper is structured as follows. The \textit{Methods} section briefly outlines the methods under investigation. In the subsequent \textit{Benchmark experiment} section we describe the conducted experiment. The findings are presented in the \textit{Results} section, which is followed by a discussion.

\section{Methods}

\subsection{Preliminary remarks}
There are essentially two ways to include multi-omics data in a prediction model. The first approach, which we term as \lq\lq naive'', does not distinguish the different data types, i.e. does not take the group structure into account. In the second approach, the group structure is taken into account. The advantage of the naive approach, its simplicity, comes at a price. First of all, physicians and researchers often have some kind of prior knowledge of which data type might be especially useful in the given context \citep{Klau2018}. If so, it is desirable to include such information by incorporating the group structure. Well-established prognostic clinical variables which are known to be beneficial for building prediction models for a specific disease are an important special case. In this situation, it may be useful to take the group structure into account during model building or even to treat clinical variables with priority. Otherwise these clinical variables might get lost within the huge amount of omics data \citep{Boulesteix2011}. To some extent, the same might be true for different kinds of omics data. If, for example, gene expression is expected to be more important than copy number variation data for the purpose of prediction, it might be useful to incorporate the distinction between these two data types into the prediction model or even to prioritize gene expression in some sense.

Other important aspects  of prediction models from the perspective of clinicians are sparsity, interpretability and transportability \citep{Klau2018}. Methods yielding models which are sparse with regard to the number of variables and number of omics types are often considered preferable from a practical perspective. Interpretation and practical application of the model to the prediction of independent data are easier with regression-based methods yielding coefficients that reflect the effects of variables on the outcome, than with machine learning algorithms \citep{Boulesteix2019}.

Finally, in addition to the prediction performances of the different methods,
the question of the additional predictive value of omics data compared to clinical data is also interesting from a clinical perspective \citep{Boulesteix2011}. 
Many of the omics-based prediction models which were claimed to be of value for predicting disease outcomes could eventually not be shown to outperform clinical models in independent studies \citep{Boulesteix2011, Bovelstad2009, BinHerold2014}.  
However, some findings suggest that using both clinical and omics variables {\it jointly} may outperform clinical models \citep{Bovelstad2009, BinSauerbrei2014, Binder2008}. In our benchmark study, we can address this issue by systematically comparing the performance of clinical models and combined models for a large number of datasets.

The methods included in our study can be subsumed in three general approaches, which are briefly described in the following subsections: penalized-regression-based methods, boosting-based methods and random-forest-based methods. A more technical description of the methods can be found in the supplementary material. Additionally, two reference methods are considered: \textit{simple Cox regression}, which only uses the clinical variables, and the \textit{Kaplan-Meier estimate}, which does not use any information from the predictor variables. 

\subsection{Penalized-regression-based methods}
The penalized regression methods briefly reviewed in this section have in common that they modify maximum partial likelihood estimation by applying a regularization, most importantly to account for the $n<<p$ problem.

Standard \textit{Lasso}, introduced more than two decades ago \citep{Tibshirani1996} and subsequently extended to survival time data  \citep{Tibshirani1997}, applies $L_1$-regularization to penalize large (absolute) coefficient values. The result is a sparse final model:  a  number of coefficients are set to zero. The number of non-zero coefficients decreases with increasing penalty parameter $\lambda$ and cannot exceed the sample size. The method does not take the group structure into account. The parameter $\lambda$ is a hyper-parameter to be tuned.

\textit{Two-step} (TS) \textit{IPF-Lasso} \citep{Schulze2017} is an extension of the standard Lasso specifically designed to take a multi-omics group structure into account. This method is an adaptation of the integrative Lasso with penalty factors (IPF) \citep{Boulesteix2017}, which consists in allowing different penalty values for each data type. In TS-IPF-Lasso, the ratios between these penalty values are determined in a first step (roughly speaking, by  applying standard Lasso and averaging the resulting coefficients).

\textit{Priority-Lasso} \citep{Klau2018} is another Lasso-based method designed for the incorporation of different  groups of variables. Often clinical researchers prioritize variables that are easier, cheaper to measure or known to be good predictors of the outcome. The principle of priority-Lasso is to  define a priority order for the groups of variables. Priority-Lasso then successively fits Lasso-regression models to these groups, whereby at each step the resulting linear predictor is used as an offset for the Lasso model fit to the next group. 
For the study at hand, however, we do not have any substantial domain knowledge, so we cannot specify a meaningful priority order. We therefore alter the method into a two-step procedure similar to the TS IPF-Lasso. More precisely, we order the groups of variables according to the mean values of their coefficients fitted in the first step by separately modeling each group. This ordering is used as a surrogate for a knowledge based priority order.

\textit{Sparse Group Lasso} (SGL) \citep{Simon2013} is another extension of the Lasso, capable of including group information. The method incorporates a convex combination of the standard Lasso penalty and the  group-Lasso penalty \citep{Yuan2006}. This simultaneously leads to sparsity on feature as well as on group level. 

\textit{Adaptive group-regularized ridge regression} (GRridge) \citep{Wiel2015} is designed to use group specific co-data, e.g., p-values known from previous studies. Multi-omics group structure may also be regarded as co-data, although the method was originally not intended for this purpose. It is based on \textit{ridge regression}, which uses a $L_2$-based penalty term. Feature selection is achieved post-hoc by exploiting the heavy-tailed distribution of the estimated coefficients, which clearly separates coefficients close to zero from those which are further away \citep{Wiel2015}. 

\subsection{Boosting-based methods}
Boosting is a general technique introduced in the context of classification in the machine learning community, which has then been revisited in a statistical context \citep{Friedman2001}. 
Statistical boosting can be seen as a form of iterative function estimation by fitting a series of weak models, so called base learners. In general, one is interested in a function that minimizes the expected loss when used to model the data. This target function is updated iteratively, with the number of boosting steps $m_{stop}$, i.e. the number of  iterations, being the main tuning parameter. This parameter, together with the so-called learning rate, which steers the contribution of each update, also leads to a feature selection property. In this study we use two different boosting approaches.

\textit{Model-based boosting} \citep{Hothorn2006}, the first variant, uses simple linear models as base learners and updates only the loss minimizing base learner per iteration. The learning rate is usually fixed to a small value such as $0.1$ \citep{Buehlmann2007}. 

\textit{Likelihood-based boosting} \citep{Tutz2006}, in contrast, uses a penalized version of the likelihood as loss and the shrinkage is directly applied in the coefficient estimation step via a penalty parameter. It is also an iterative procedure: the updates of previous iterations are included as an offset to make use of the information gained.

\subsection{Random-forest-based methods}
\textit{Random forest} is a tree-based ensemble method introduced by \cite{Breiman2001}. Instead of growing a single classification or regression tree, it uses bootstrap aggregation to grow several trees and averages the results. Random forest was later expanded to survival time data \citep{Ishwaran2008}. For each split in each tree, the variable maximizing the difference in survival is chosen as best feature. Eventually, the cumulative hazard function is computed via the Nelson-Aalen estimator in each final node in each tree. For prediction, these estimates are averaged across the trees to obtain the ensemble cumulative hazard function.

\textit{Block forest} \citep{Hornung2018} is a variant modifying the split point selection of random forest to incorporate the group  structure (or \lq\lq block'' structure, hence the name of the method) of multi-omics data. It can be applied to any outcome type for which a random forest variant exists. 

\section{Benchmark experiment}

\subsection{Study design}
Our study is intended as a {\it neutral comparison study}; see \citet{BoulesteixLauer2013,BoulesteixWilson2017} for an exact definition and discussions of this concept. Firstly, we compare methods that have been described elsewhere and do not aim at emphasizing a particular method. Secondly, we tried to achieve a reasonable level of neutrality, which we disclose here following the example of \citet{Couronne2018}. As a team we are approximately equally familiar with all classes of methods. Some of us have been involved in the development of priority-Lasso, IPF-Lasso and block forest. As far as the other methods are concerned, we contacted the person listed in CRAN as package maintainer via email and asked for an evaluation of our implementation including the choice of parameters.

A further important aspect of the study design is the choice and number of datasets used for the comparison, since the performance of prediction methods usually strongly varies across datasets.
\citet{BoulesteixWilson2017} compare benchmark experiments to clinical trials, where methods play the role of treatments and datasets play the role of patients. In analogy to clinical trials, the number of considered datasets should be chosen large enough to draw reliable conclusions, and the selection of datasets should follow strict inclusion criteria and not be modified after seeing the results; see the \textit{Datasets} section for more details on this process. 
Finally, a benchmark experiment should be easily extendable (and, of course, reproducible). It is almost impossible to include every available method in a single benchmark experiment, and it should also be easy to compare methods proposed later without re-running the full experiment and without too much programming effort. For this reason we use the R package \textit{mlr} \citep{mlr}, which offers a unified framework for benchmark experiments and makes them easily extendable and reproducible.

\subsection{Technicalities and implementation}
The benchmark experiment is conducted using R 3.5.1 \citep{RCore2014}. We compare the $14$ learners described in the \textit{Method configurations} section on $18$ datasets (see the \textit{Datasets} section). 
The code to reproduce the results is available on GitHub\footnote{\url{https://github.com/HerrMo/multi-omics_benchmark_study}}, the data can be obtained from OpenML\footnote{\url{https://www.openml.org/}} \citep{OpenML2013, OpenMLR2017}. To further improve reproducibility, the package \textit{checkpoint} \citep{checkpoint} is used. Because the computations are time demanding but parallelisable, the package \textit{batchtools} \citep{batchtools} is used for parallelisation. The package \textit{mlr} \citep{mlr}, used for this benchmark experiment, offers a simple framework to conduct all necessary steps in a unified way. \\
We use 10 x 5-fold CV for datasets with a size less than 92 MB (11 datasets) and 5 x 5-fold CV for datasets with a size larger than 112 MB (7 datasets) to keep computation times feasible. Furthermore, we stratify the subsets, since the ratio of events and censorings is unbalanced for some datasets. Moreover, hyperparameter tuning is performed. This could in principle also be implemented via \textit{mlr}, but in this study the tuning procedures provided by the specific packages are used. We denote the resampling strategy used for hyper-parameter tuning inner-resampling and the repeated CV used for performance assessment outer-resampling. For inner-resampling we use out-of-bag-samples (OOB) for random forest learners and 10-fold CV for the other learners. 

\subsection{Performance evaluation}
The performance is evaluated in three dimensions. First of all, the prediction performance is assessed via the integrated Brier score and the C-index suggested by \cite{Uno2011} (hereinafter simply denoted as ibrier and cindex). \\
The second dimension is the sparsity of the resulting models, which has two aspects: sparsity on the level of variables and sparsity on group level. The latter refers to whether variables of only some groups are selected. Sparsity on feature level, in contrast, refers to the overall sparsity, i.e. the total number of selected features. As random forest does not perform variable selection, it is not assessed in this dimension. Computation times are considered as a third dimension. 

Another important aspect is the different use of group structure information. Some of the methods do not use any such information, some favor clinical data over molecular data, and some differentiate between all groups of variables (i.e., also between omics groups). Thus, the differences in performance might not only result from using different prediction methods. They may also arise from the way in which the group structure information is included. Therefore, comparability in terms of predictive performance is only given for methods that use the same strategy to include group information: (1) \textit{naive methods} not using the group structure, (2) methods using the group structure and \textit{not favoring clinical features}, (3) methods using the group structure by \textit{favoring clinical features}, where we subsume methods favoring clinical and not distinguishing molecular covariates, and methods favoring clinical and additionally also distinguishing molecular covariates. 

\subsection{Method configurations}
Following the terminology of the package \textit{mlr} \citep{mlr}, we denote a method configuration as a  \lq\lq learner''. There may be several learners based on the same method. An overview of learners considered in our benchmark study is displayed in Table \ref{tab:lrns}, while the full specification is given in the paragraph devoted to the corresponding method. In the following, the R packages used to implement the learners can be found in parentheses after the paragraph heading.   \\
\\
\textit{Penalized-regression-based learners}\\
\\
\textit{Lasso} (glmnet \cite{glmnet, Simon2011})\\
The penalty parameter $\lambda$ is chosen via internal 10-fold CV. No group structure information is used.\\
\\
\textit{SGL} (SGL \cite{SGL}) \\
The model is fit via the \textit{cvSGL} function. Tuning of the penalty parameter $\lambda$ is conducted via internal 10-fold CV. The parameter $\alpha$ steering the contribution of the group-Lasso and the standard Lasso is not tuned and set to the default value 0.95, as recommended by the authors \citep{Simon2013}. All other parameters are set to default as well.\\
\\
\\
\textit{TS IPF-Lasso} (ipflasso \cite{ipflasso}) \\
The penalty factors are selected in the first step by computing separate ridge regression models for every feature group and averaging the coefficients within the groups by the arithmetic mean. These settings have shown reasonable results \citep{Schulze2017}. The choice of the penalty parameters $\lambda_m$ is conducted using 5-fold-CV in the first step and 10-fold CV in the second.\\
\\
\textit{priority-Lasso} (prioritylasso \cite{prioritylasso})\\
The priority order is determined through a preliminary step realized in the same way as in the first step of TS IPF-Lasso.   The priority-Lasso method takes into account the group structure. Even though the version with cross-validated offsets delivers slightly better prediction results \citep{Klau2018}, the offsets are not estimated via CV in order to not increase the computation times further. To select the parameter $\lambda$ in each step of priority-Lasso, 10-fold CV is used. \\
\\
\textit{priority-Lasso favoring clinical features} (prioritylasso \cite{prioritylasso}) \\
The settings are the same as before, except that the group of clinical variables is always assigned the highest priority. The preliminary step only determines the priority order for the molecular groups. The clinical variables are used as an offset when fitting the model of the second group. Furthermore, the clinical variables are not penalized (setting parameter  block1.penalization = FALSE). \\
\\
\textit{GRridge} (GRridge \cite{ggridge}) \\
This method was not originally intended for the purpose of including multi-omics group structure, but is capable of doing so. To better fit the task at hand, a special routine was provided by the package author in personal communication. 
In addition, the argument \textit{selectionEN} is set to \textit{TRUE} so post-hoc variable selection is conducted, and \textit{maxsel}, the maximum number of variables to be selected, is set to 1000.\\
\\
\noindent \textit{Boosting-based learners}\\
\\
\textit{Model-based boosting} (mboost \cite{mboost})\\
Internally, \textit{mlr} uses the function \textit{glmboost} from the package \textit{mboost} and sets the family argument to \textit{CoxPH()}. Furthermore the  number of boosting steps ($m_{stop}$) is chosen by a 10-fold CV on a grid from 1 to 1000 via \textit{cvrisk}. For the learning rate $\nu$ the default value of 0.1 is used. Group structure information is not taken into account.\\
\\
\textit{Likelihood-based boosting} (CoxBoost \cite{CoxBoost}) \\ 
The maximum number of boosting steps \textit{maxstepno} is set to default, i.e. 100. Again, $m_{stop}$ is determined by 10-fold CV. The penalty $\lambda$ is set to default and thus computed according to the number of events. No group structure information is used.\\
\\
\textit{Likelihood-based boosting favoring clinical features} (CoxBoost \cite{CoxBoost}) \\
The settings are the same as before. Additionally, group structure information is used by specifying the clinical features as mandatory. These features are favored as in the case of priority-Lasso by setting them as an offset and not penalising them. Further group information is not used, so the molecular data are not distinguished.\\
\\
\textit{Random-forest-based learners}\\
\\
\textit{Random forest} (randomForestSRC \cite{RFSRC}; ranger \cite{ranger})\\
Two versions of standard \textit{random forest} are examined: \textit{randomForestSRC (rfsrc)} and \textit{ranger}. They share the same theoretical background, but differ in their implementation. Thus, two different implementations are compared. Tuning of \textit{mtry} is conducted via the \textit{tune} function for \textit{rfsrc} and via the \textit{tuneMtryfast} function of package \textit{tuneRanger} for \textit{ranger}. For \textit{tune} the default settings have been altered, firstly to match \textit{tuneMtryfast} and secondly because they lead to infeasible computation times. The minimal node size is 3 for both (the \textit{ranger} default settings). The other hyper-parameters are set to defaults. \\
\\
\\
\\
\textit{Block forest} (blockForest \cite{blockForest}) \\
Block forest is a random forest variant able to include group structure information. The implementation is based on \textit{ranger}. With function \textit{blockfor} the models are fit via the default settings. \\
\\
\textit{Reference methods}\\
\\
The clinical reference model is a Cox proportional hazard model, computed via the \textit{coxph} function of the \textit{survival} package \citep{survival} and only uses clinical features. The Kaplan-Meier estimate is computed via \textit{survfit} from the same package.

\subsection{Datasets}
\label{subsec:datasets}
From the cancer datasets that have been gathered by the TCGA research network\footnote{\url{https://cancergenome.nih.gov}}, we selected those with more than 100 samples and five different multi-omics groups, which resulted in a collection of datasets for 26 cancer types. As described below, further preprocessing eventually lead to 18 usable datasets. 
Table \ref{tab:datsets} lists all cancer types and the abbreviations used to reference them within the study.

For each cancer type there are four molecular data types and the clinical data type, i.e. five groups of variables. The molecular data types comprise copy number variation (cnv), gene expression (rna), miRNA expression (mirna), and mutation. The number of variables differs strongly between groups but is similar across datasets. Most molecular features (about 60,000) belong to the cnv group but only a few hundred features to mirna, the smallest group. There is a total of about 80,000 to 103,000 molecular features for each cancer type.

Of the 26 available datasets, three were excluded because they did not have observations for every data type. Furthermore, since the outcome of interest is survival time, not only the number of observations is crucial but, most importantly, the number of events (deaths), which we call the number of effective cases. A ratio of 0.2 of effective cases is common \citep{BinSauerbrei2014}. The five datasets that had less than 5~\% effective cases were excluded.

Since the majority of the clinical variables had missing values, the question arose of which to include for a specific dataset while saving as many observations as possible. As we did not have any domain knowledge, we adopted  a two-step strategy. 
Firstly, an informal literature search was conducted to find studies where the specific cancer type was under investigation. Variables mentioned to be useful in these studies were included, if available. Secondly, we additionally used variables that were available for most of the cancer types. These comprised sex, age, histological type and tumor stage. These were included as standard, if available. Of course, sex was not included for the sex-specific cancer types. 

\begin{table*}[!h]
 \caption{Summary of learners used for the benchmark experiment. Use of group structure information is indicated by *.\label{tab:lrns}}
    \centering
    {\begin{tabular}{@{}llll@{}}
        \toprule
        learner & method & package::function & tuning\\
        \midrule
        Lasso & Standard Lasso & glmnet::cv.glmnet & 10-f-CV\\
        ipflasso* & TS IPF-Lasso & ipflasso::cvr.ipflasso & 10-f-CV\\
        prioritylasso* & priority-Lasso & priortiylasso::prioritylasso & 10-f-CV\\
        prioritylasso favoring* & priority-Lasso & priortiylasso::prioritylasso & 10-f-CV\\
        grridge* & GRridge & GRridge::grridge & 10-f-CV\\
        SGL* & SGL & SGL::cvSGL & 10-f-CV\\
        glmboost & Model-based boosting & mboost::glmboost & 10-f-CV\\
        CoxBoost & Likelihood-based boosting & CoxBoost::cv.CoxBoost & 10-f-CV\\
        CoxBoost favoring* & Likelihood-based boosting & CoxBoost::cv.CoxBoost & 10-f-CV\\
        rfsrc & Random forest & randomForestSRC::tune & OOB\\
        ranger & Random forest & tuneRanger::tuneMtryfast & OOB\\
        blockForest* & Block forest & blockForest::blockfor & OOB\\
        Clinical only & Cox model & survival::coxph & no\\
        Kaplan-Meier & Kaplan-Meier estimate & survival::survival & no\\
        \bottomrule
    \end{tabular}}
\end{table*}

\begin{table*}[!h]
    \caption{Summary of the datasets used for the benchmark experiment. The third to the seventh column show the number of features in the feature group, the eighth column the total amount of features (p). The last three columns show, in this order, the number of observations (n), the number of effective cases (n\_e) and the ratio of the number of events and the number of observations (r\_e). Abbreviations: C. = Carcinoma, CC. = Cell Carcinoma, PP = Renal Papilla, AC. = Adenocarcinoma, M. = Melanoma, EC. = Endometrial Carcinoma.\label{tab:datsets}}
    \centering
    \resizebox{\linewidth}{!}
    {\begin{tabular}{@{}lllllllllll@{}}
    		\toprule
        dataset & cancer & clin. & cnv & mirna & mutation & rna & p & n & n\_e & r\_e\\
        \midrule
        BLCA & Bladder Urothelial & 5 & 57964 & 825 & 18577 & 23081 & 100455 & 382 & 103 & 0.27\\
        BRCA & Breast Invasive C. & 8 & 57964 & 835 & 17975 & 22694 & 99479 & 735 & 72 & 0.10\\
        COAD & Colon Adenocarcinoma & 7 & 57964 & 802 & 18538 & 22210 & 99524 & 191 & 17 & 0.09\\
        ESCA & Esophageal C. & 6 & 57964 & 763 & 12628 & 25494 & 96858 & 106 & 37 & 0.35\\
        HNSC & Head-Neck Squamous CC. & 11 & 57964 & 793 & 17248 & 21520 & 97539 & 443 & 152 & 0.34\\
        KIRC & Kidney Renal Clear CC. & 9 & 57964 & 725 & 10392 & 22972 & 92065 & 249 & 62 & 0.25\\
        KIRP & Cervical Kidney RP. CC. & 6 & 57964 & 593 & 8312 & 32525 & 99403 & 167 & 20 & 0.12\\
        LAML & Acute Myeloid Leukemia & 7 & 57962 & 882 & 2176 & 29132 & 90162 & 35 & 14 & 0.40\\
        LGG & Low Grade Glioma & 10 & 57964 & 645 & 9235 & 22297 & 90154 & 419 & 77 & 0.18\\
        LIHC & Liver Hepatocellular C. & 11 & 57964 & 776 & 11821 & 20994 & 91569 & 159 & 35 & 0.22\\
        LUAD & Lung Adenocarcinoma & 9 & 57964 & 799 & 18388 & 23681 & 100844 & 426 & 101 & 0.24\\
        LUSC & Lung Squamous CC. & 9 & 57964 & 895 & 18500 & 23524 & 100895 & 418 & 132 & 0.32\\
        OV & Ovarian Cancer & 6 & 57447 & 975 & 13298 & 24508 & 96237 & 219 & 109 & 0.50\\
        PAAD & Pancreatic AC. & 10 & 57964 & 612 & 12392 & 22348 & 93329 & 124 & 52 & 0.42\\
        SARC & Sarcoma & 11 & 57964 & 778 & 10001 & 22842 & 91599 & 126 & 38 & 0.30\\
        SKCM & Skin Cutaneous M. & 9 & 57964 & 1002 & 18593 & 22248 & 99819 & 249 & 87 & 0.35\\
        STAD & Stomach AC. & 7 & 57964 & 787 & 18581 & 26027 & 103369 & 295 & 62 & 0.21\\
        UCEC & Uterine Corpus EC. & 11 & 57447 & 866 & 21053 & 23978 & 103358 & 405 & 38 & 0.09\\
        \bottomrule
    \end{tabular}}
\end{table*}

\section{Results}
\subsection{Failures and refinement of the study design}
For some CV iterations the model fitting was not successful, leading to NAs for the assessment measures for these iterations. This is common in benchmark experiments of larger scale \citep{BischlSchiffner2013}. To cope with such modeling failures we follow strategies described previously \citep{BischlSchiffner2013,Probst2019}. If a learner fails in more than 20~\% of the CV iterations for a given dataset, we assign (for the failing iterations) values of the performance measures corresponding to random prediction (0.25 for ibrier and 0.5 for cindex) and the mean of the other iterations for the computation time and the number of selected features. If a learner fails in less than 20~\%, the performance means of the successful iterations are assigned for all measures.

Besides such modeling failures, for two methods more general  issues related to usability occurred while conducting the experiment. First of all, not all methods could be run on the same system. For the majority of the methods we used the Linux distribution Ubuntu 14.04, but using rfsrc together with batchtools under Ubuntu was not possible, so it was run using Windows 7. If no parallelization is applied via batchtools, rfsrc also works using Ubuntu.

In contrast, SGL did not work using Windows 7, but worked using Ubuntu irrespective of the parallelization issue. Using SGL with the considered large numbers of features always leads to a fatal error in R under Windows. 
The extremely long computation times for SGL were problematic. Since we received no feedback from the authors, we used the standard settings. These lead to computations lasting several days for one single model fit for large datasets. Altering some of the parameters did not strongly reduce the computational burden. Running the whole experiment as planned was thus not possible for SGL. Here we briefly present the results of SGL which could be obtained based on four of the smallest datasets. For the rest of the study we exclude SGL from the analysis. On average over all iterations and the four datasets, SGL leads to a cindex of 0.58 and an ibrier of 0.24. The resulting models are neither sparse on feature level, with an average of 4083 selected features, nor on group level. The mean computation time of 5.7 hours for one CV-iteration confirms the problem of extremely long computation times. In comparison, the next slowest method needs 1.2 hours for one iteration, on average over all datasets. Table \ref{tab:SGL} shows the performance values of SGL for each of the four datasets and the mean values.
\begin{table}[!ht]
    \caption{Performance of SGL on four small datasets. Column \lq\lq all'' represents the total number of selected features, the subsequent columns the numbers of selected features of the respective groups.\label{tab:SGL}}
    \centering
    {\begin{tabular}{@{}llllllllll@{}} 
    		\toprule
        data & cindex & ibrier & time & all & clin & cnv & mirna & rna & mut \\
        \midrule
        LAML & 0.496 & 0.231 & 1.9 & 8149 & 0.5 & 7822 & 4.7 & 0 & 323 \\
        LIHC & 0.533 & 0.198 & 9.0 & 3617 & 0.3 & 3250 & 28  & 264 & 75 \\
        PAAD & 0.650 & 0.255 & 4.5 & 1483 & 3.2 & 62   & 30  & 12  & 1375 \\
        SARC & 0.629 & 0.278 & 7.5 & 3081 & 2.7 & 1906 & 51  & 40  & 1082 \\ 
        mean & 0.58 & 0.24 & 5.7 & 4083 & 1.7 & 3260 & 28 & 79 & 714 \\ 
        \bottomrule
    \end{tabular}}{}
\end{table}

\subsection{Computation time}
Table \ref{tab:ave} shows the average performance measures for every method and is ordered by the cindex. All values are obtained by---firstly---averaging over the outer-resampling CV-iterations and---secondly---averaging over the datasets. For the methods not yielding model coefficients the corresponding cells contains \lq\lq-''.

The eighth column of Table \ref{tab:ave} displays the mean computation time. The computation times are measured as the time needed for model fitting (training time).
The fastest procedures are standard \textit{Lasso} and \textit{ranger}, followed by \textit{glmboost} and the \textit{CoxBoost} variants. The three penalized regression methods using group structure (IPF-Lasso, priority-Lasso, GRridge) are about 2 to 3 times slower, with GRridge being the fastest of the three methods. \textit{rfsrc} needs about as much time as \textit{prioritylasso} and \textit{ipflasso}. Of the two \textit{prioritylasso} variants the one favoring clinical features is a little slower. Finally, \textit{blockForest} is the slowest method.

\begin{figure*}[!ht]
    \centering
    \includegraphics[height=12.5cm]{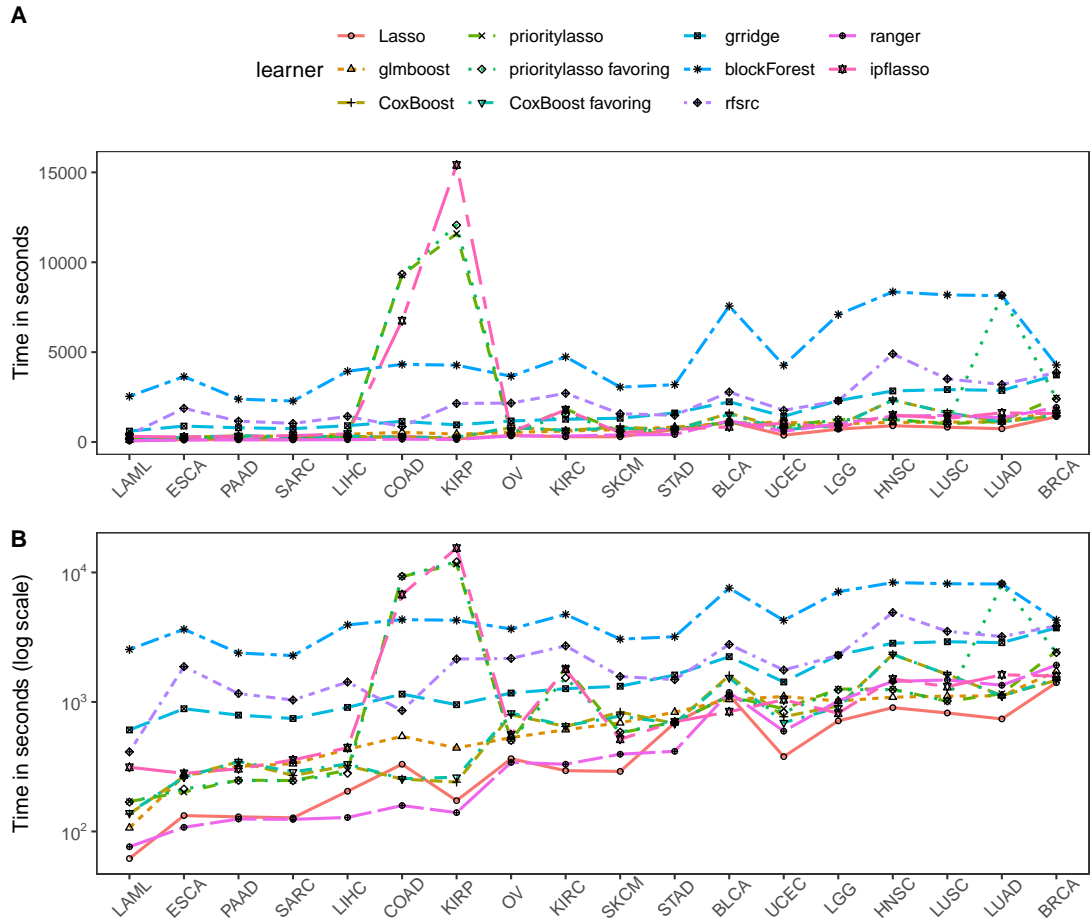}
    \caption{\label{fig:times}Computation time. (A) Computation times in seconds. (B) Computation times in log(seconds). The datasets are ordered form smallest (LAML) to largest (BRCA).}
\end{figure*}

Of course, the computation times depend on the size of the datasets. Figure \ref{fig:times} displays the mean computation time of one outer-resampling iteration for the different learners and datasets. The datasets are ordered from smallest (LAML) to largest (BRCA).  
It can also be seen that the effective number of cases influences the computation time. COAD and KIRP are among the smaller datasets with 17 (9 \%) and 20 (12 \%) events, respectively. IPF-Lasso and priority-Lasso yield a substantial increase in computation time for these datasets. 

\begin{figure*}[!h]
    \centering
    \includegraphics[height=8cm]{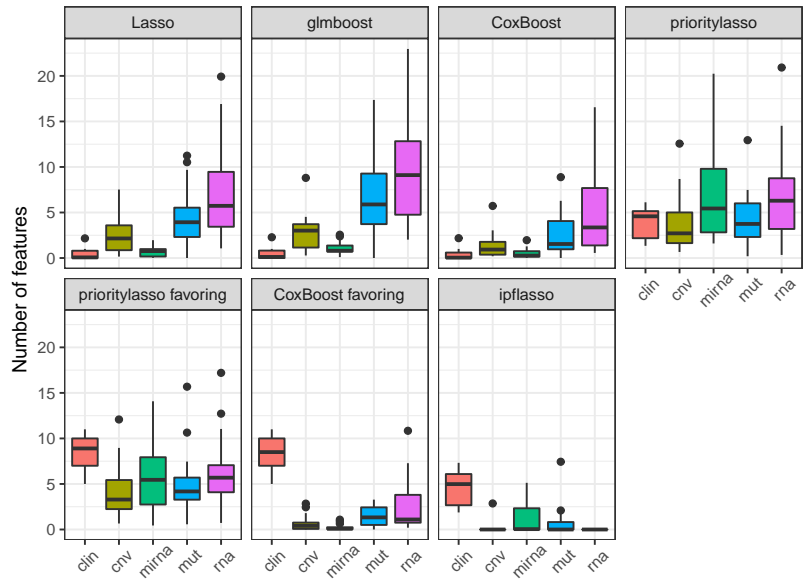}
    \caption{\label{fig:spars}Sparsity on group level. Only learners yielding model coefficients are considered. Each panel shows the number of selected features for every feature group for the considered learners. grridge is excluded since it yields models on a much larger scale.}
\end{figure*}

\subsection{Model sparsity}
To assess sparsity, the number of non-zero coefficients of the resulting model of each CV iteration is considered. As random forest models do not yield such coefficients, this aspect is not assessed for random forest variants.

\subsubsection{Sparsity on the level of variables}
Sparsity in terms of the number of included variables is particularly interesting for practical purposes, since sparse models are easier to interpret and to communicate. On average, as Table \ref{tab:ave} shows, \textit{ipflasso} leads to the sparsest models with an average of 7 variables, followed by \textit{CoxBoost} with on average 10 variables. \textit{CoxBoost favoring} and \textit{Lasso} are also reasonably sparse (13, 16), but the variability is higher for \textit{Lasso} (see Figure \ref{fig:perf}). \textit{glmboost}, \textit{prioritylasso} and \textit{prioritylasso favoring} yield models with more than 20 features (22, 26, 30). Least sparse is \textit{grridge}; the average \textit{grridge} model size (984) is close to the maximum number of features to be selected (\textit{maxsel = 1000}). \textit{grridge} seems not to be able to appropriately select variables in this setting (recall that it is  not intended to do so).

\subsubsection{Sparsity on group level}
Figure \ref{fig:spars} displays the number of selected variables by group for all learners except \textit{grridge} (since \textit{grridge} leads to models on a much larger scale), the random forest variants and the reference learners. Table \ref{tab:ave} shows that \textit{grridge} chooses variables from all groups and is thus not sparse on group level.

Among the other methods, IPF-Lasso yields strong sparsity on group level. Mostly clinical features are selected. The priority-Lasso is not able to eliminate groups. 
Furthermore, with boosting variants \textit{CoxBoost} and \textit{glmboost} and with standard \textit{Lasso} no clinical features are selected. \textit{CoxBoost favoring} does not select mirna features. IPF-Lasso does not include cnv and rna features.
This exemplifies the problem of methods treating high- and low-dimensional groups equally. As already pointed out, due to their low dimension, clinical variables get lost within the huge number of molecular variables. It becomes obvious that this also applies for some of the molecular variables. The mirna group is, in comparison to the other molecular groups, lower dimensional with 585 to 1002 features. Learners which do not consider group structure fail to include clinical variables and include at most one mirna variable.
\textit{CoxBoost favoring}, which differentiates clinical and molecular variables, does not select mirna features. 
In contrast, learners taking into account the multi-omics group structure generally include variables of both lower dimensional groups. Using the group structure thus prevents low-dimensional groups from being discounted. 

\subsection{Prediction performance}
\subsubsection{Overview and main findings}
\begin{figure*}[!h]
    \centering
    \vspace{1cm}
    \includegraphics[height=11cm]{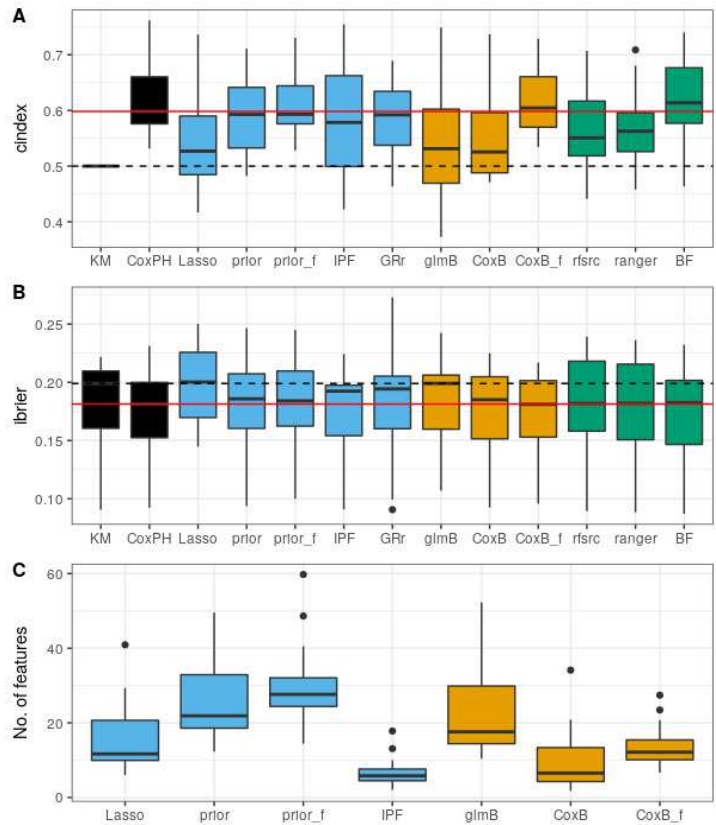}
    \caption{\label{fig:perf}Performance  of the learners - A: cindex. B: ibrier. C: Total number of selected features; only learners yielding model coefficients are included and grridge is excluded since it yields models on a much larger scale. The solid red and dashed black horizontal lines correspond to the median performance of the clinical-only-model and the Kaplan-Meier-estimate, respectively. Colours indicate membership to one of the general modeling approaches: penalised regression (blue), boosting (orange), random forest (green), reference methods (black). Abbreviations: KM = Kaplan-Meier, Lasso = Lasso, glmB = glmboost, CoxB = CoxBoost, CoxPH = Clinical only, prior = prioritylasso, prior\_f = prioritylasso favoring, IPF = ipflasso, CoxB\_f = CoxB favoring, GRr = grridge, BF = blockForest, rfsrc = rfsrc, ranger = ranger.}
\end{figure*}

Figure \ref{fig:perf} shows the distributions of the values of the performance metrics across the datasets. Again, \textit{grridge} is excluded from the sparsity panel.
Three important findings can be highlighted. First of all, regarding Figure \ref{fig:perf}, most of the learners perform better than the Kaplan-Meier estimate (indicated by the dashed horizontal line). This indicates that using the variables is, in general, useful. Only \textit{Lasso} performs worse than the Kaplan-Meier-estimate (based on the ibrier). 
Secondly, only \textit{blockForest} outperforms the reference clinical Cox model  (red horizontal line), which stresses the importance of the clinical variables for these datasets.
Finally, methods taking into account the group structure in some way in general outperform the naive methods according to the cindex. 

\subsubsection{Comparing prediction methods}
All analyses in this section refer to Table \ref{tab:ave}. 

\paragraph{Naive methods:} 
The learners \textit{CoxBoost}, \textit{glmboost}, \textit{rfsrc}, \textit{ranger} and standard \textit{Lasso} are fit with the naive strategy. 
In general, the results are not consistent over the two measures. According to the cindex, random forest performs best, regardless of which implementation is used, followed by likelihood-based boosting and Lasso. Model-based boosting performs worst. All methods are at least slightly better than the Kaplan-Meier estimate. With regards to the ibrier, however, likelihood-based boosting performs best. Moreover, model-based boosting performs better than Lasso, but still gets outperformed by the random forest variants which are close to likelihood-based boosting. To sum up, although the results differ depending on the considered measure, random forest shows a tendency to outperform the other methods, since it performs best based on the cindex and is among the best methods based on the ibrier.

\paragraph{Methods not favoring clinical features:} The learners \textit{block forest}, \textit{ipflasso},  \textit{prioritylasso}, and \textit{grridge} use the group structure but do not favor clinical features.  The random forest variant \textit{blockForest} outperforms the other methods. It performs better on average than any other method based on both measures. Among the penalized regression methods, priority-Lasso performs best according to the cindex and IPF-Lasso according to the ibrier. GRridge ranks second according to the cindex and third according to the ibrier. 
Moreover, priority-Lasso and GRridge perform equal to or even worse than the Kaplan-Meier estimate based on the ibrier. Since IPF-Lasso yields the sparsest models, it might be preferable when sparsity is important.

\paragraph{Methods favoring clinical features:}
There are two learners favoring clinical features: \textit{CoxBoost favoring} and \textit{prioritylasso favoring}. 
The results are unambiguous with \textit{CoxBoost favoring} performing better than \textit{prioritylasso favoring}. Furthermore, both learners perform better than the Kaplan-Meier estimate based on the cindex, but only \textit{CoxBoost favoring} performs better than the Kaplan-Meier estimate based on the ibrier. 
Thus, according to these findings, likelihood-based boosting yields better results than priority-Lasso when clinical variables are favored, even though priority-Lasso here further distinguishes the molecular data. \\

\begin{table*}[!h]
    \caption{\label{tab:ave}Average learner performances. The values are obtained by averaging over the CV-Iterations and datasets. The time is measured in minutes. For learners not yielding model coefficients, the corresponding measures are set to \lq\lq-''. Column \lq\lq all'' represents the total number of selected features, the subsequent columns the numbers of selected features of the respective groups. The \lq\lq ci'' columns display the 95~\% confidence intervals for the means based on quantiles of the t-distribution. The total number of features may differ from the sum of features in each group due to rounding errors.}
    \centering
    \resizebox{\linewidth}{!}
    {\begin{tabular}{@{}llllllllllllll@{}}
    \toprule
     & & cindex &  &  & ibrier &    &      &     &      &     &       &     &    \\
	learner & mean   & sd & ci & mean & sd & ci & time & all & clin & cnv & mirna & rna & mut\\
	\midrule
	blockForest & 0.620 & 0.072 & [0.584, 0.656] & 0.174 & 0.042 & [0.153, 0.195] & 80 & - & - & - & - & - & -\\
	Clinical only & 0.618 & 0.060 & [0.588, 0.648] & 0.175 & 0.038 & [0.156, 0.194] & 0 & 8 & 8 & - & - & - & -\\
	CoxBoost favoring & 0.618 & 0.057 & [0.589, 0.646] & 0.174 & 0.036 & [0.156, 0.192] & 13 & 13 & 8 & 1 & 0 & 3 & 1\\
	prioritylasso favoring & 0.607 & 0.056 & [0.579, 0.635] & 0.181 & 0.040 & [0.161, 0.201] & 39 & 30 & 8 & 4 & 6 & 6 & 5\\
	prioritylasso & 0.591 & 0.068 & [0.558, 0.625] & 0.180 & 0.037 & [0.162, 0.199] & 32 & 26 & 4 & 4 & 7 & 7 & 4\\
	grridge & 0.587 & 0.069 & [0.553, 0.622] & 0.181 & 0.044 & [0.159, 0.203] & 28 & 984 & 2 & 332 & 19 & 540 & 91\\
	ipflasso & 0.578 & 0.100 & [0.528, 0.628] & 0.176 & 0.034 & [0.159, 0.193] & 33 & 7 & 5 & 0 & 1 & 0 & 1\\
	rfsrc & 0.567 & 0.078 & [0.528, 0.606] & 0.182 & 0.045 & [0.160, 0.204] & 36 & - & - & - & - & - & -\\
	ranger & 0.562 & 0.068 & [0.529, 0.596] & 0.179 & 0.045 & [0.157, 0.202] & 10 & - & - & - & - & - & -\\
	CoxBoost & 0.552 & 0.080 & [0.512, 0.592] & 0.175 & 0.039 & [0.155, 0.194] & 14 & 10 & 0 & 1 & 1 & 5 & 2\\
	Lasso & 0.546 & 0.089 & [0.502, 0.591] & 0.198 & 0.034 & [0.181, 0.215] & 8 & 16 & 0 & 2 & 1 & 8 & 5\\
	glmboost & 0.542 & 0.104 & [0.490, 0.593] & 0.188 & 0.037 & [0.169, 0.206] & 12 & 22 & 0 & 3 & 1 & 10 & 7\\
	Kaplan-Meier & 0.500 & 0.000 & [0.5, 0.5] & 0.180 & 0.040 & [0.160, 0.200] & 0 & - & - & - & - & - & -\\
	\bottomrule
    \end{tabular}}{}
\end{table*}

However, when comparing the described performances, one has to take into account the small differences in performance and the variability of the method performances across datasets. As the confidence intervals in Table \ref{tab:ave} show, conclusions about the superiority of one method over another should be treated with caution. Moreover, paired two-sided t-tests comparing, for example, \textit{blockForest} with \textit{CoxBoost favoring} and the \textit{Clinical only} model show no statistically significant differences in performance with p-values of 0.81 and 0.86 (cindex) and 0.95 and 0.78 (ibrier). For all t-tests in the study, the normal distribution assumption was checked with Shapiro-Wilk-tests and Q-Q-plots. 

\subsubsection{Using multi-omics data}
\paragraph{Added predictive value:}
To assess the added predictive value of the molecular data, we follow approach A proposed by \citet{Boulesteix2011}, thus comparing learners obtained by only using clinical features and combined learners, i.e. learners using clinical and molecular variables. Since it is emphasized that for this validation approach the combined learners should not be derived by the naive strategy, these learners are not considered here. 

In general, the findings indicate that the multi-omics data may have the potential to add predictive value. First of all, \textit{blockForest} outperforms the Cox model based on both measures. Secondly, as Table \ref{tab:group} shows, there are several datasets for which there is at least one learner that takes the group structure into account and outperforms the clinical learner. For some of the datasets, e.g. LAML and COAD, the performance differences are substantial. Thus, using additional molecular data leads to better prediction performances in some of the considered cases. On the other hand, it must take into account that in the other cases the differences are small and, again regarding the confidence intervals, one has to be careful when drawing conclusions about the superiority over the Cox model. Moreover, for six datasets the Cox model does not get outperformed at all by methods which use the omics data. This raises serious concerns regarding a beneficial effect of the omics data in general. 

\paragraph{Including group structure:}
In general, the results suggest that using the naive strategy of treating clinical and molecular variables equally leads to a worse performance in comparison to methods that take the group structure into account.
Table \ref{tab:struc} shows the mean performance of the naive learners and the structured learners (both favoring and not favoring the clinical features) by dataset. 
Each value is computed as average over the naive respectively the structured learner's mean cindex and ibrier values. 

Only in five cases is the average performance of the naive learners better than the average performance of the structured learners:
Regarding the ibrier, the naive learners perform better than the structured learners for four datasets; regarding the cindex, only for the LGG dataset is the performance of the naive learners higher than the performance of the structured learners.
Unpaired, one-sided t-tests for the five naive and the six structured learners, using the mean performance values of the individual methods over the datasets as observations, yield p-values of 0.0002 and 0.0904 for the cindex and ibrier.

\paragraph{Favoring clinical features:}
According to our findings, favoring clinical variables leads to better prediction results. For likelihood-based boosting, this is in line with the findings of others (see \citet{Bin2016} and the reference therein). Differentiating the clinical variables from the molecular features strongly increases the prediction performance of likelihood-based boosting (\textit{CoxBoost} and \textit{CoxBoost favoring}), according to the average cindex. Favoring clinical features raises likelihood-based boosting from one of the worst to one of the best performing methods. Moreover, our findings show this might also hold for methods which use the multi-omics group structure. For priority-Lasso the increase is not as strong, but still notable when considering the cindex. Yet, the ibrier does not confirm this. 

\begin{table*}[!h]
\vspace{1cm}
    \caption{Assessment of the added predictive value of the molecular variables by dataset. The second and seventh column show the best performing learners for the respective dataset and measure, the \lq\lq cindex'' and \lq\lq ibrier'' column the performances of these learners. In the cases where the clinical only model is outperformed, the \lq\lq ref.'' columns show the corresponding cindex and ibrier values of the reference Cox model only using clinical variables. The \lq\lq ci'' columns show the 95~\% confidence intervals for the respective performance values based on quantiles of the t-distribution. Bold numbers indicate datasets for which there is at least one method using the group structure and outperforming the Cox model regarding both measures. \label{tab:group}}
    \centering
    \resizebox{\linewidth}{!}
    {\begin{tabular}{llllll|lllll}
    \toprule
        data & learner & cindex & ci & ref. & ci & learner & ibrier & ci & ref. & ci\\
        \midrule
BLCA & CoxBoost favoring & \textbf{0.640} & [0.612, 0.668] & 0.633 & [0.607, 0.659] & CoxBoost favoring & \textbf{0.19} & [0.181, 0.199] & 0.192 & [0.183, 0.201]\\
BRCA & CoxBoost favoring & \textbf{0.643} & [0.618, 0.669] & 0.637 & [0.608, 0.666] & blockForest & \textbf{0.141} & [0.134, 0.149] & 0.147 & [0.137, 0.158]\\
COAD & blockForest & \textbf{0.656} & [0.586, 0.725] & 0.541 & [0.475, 0.608] & blockForest & \textbf{0.087} & [0.075, 0.099] & 0.101 & [0.088, 0.115]\\
ESCA & Clinical only & 0.574 & [0.536, 0.612] & - & - & ipflasso & 0.209 & [0.198, 0.221] & 0.214 & [0.199, 0.228]\\
HNSC & blockForest & \textbf{0.582} & [0.554, 0.610] & 0.554 & [0.519, 0.588] & glmboost & \textbf{0.202} & [0.193, 0.211] & 0.210 & [0.201, 0.220]\\
KIRC & Clinical only & 0.761 & [0.734, 0.789] & - & - & ipflasso & 0.144 & [0.138, 0.149] & 0.146 & [0.140, 0.152]\\
KIRP & rfsrc & \textbf{0.648} & [0.582, 0.713] & 0.572 & [0.502, 0.641] & ranger & \textbf{0.118} & [0.106, 0.131] & 0.140 & [0.117, 0.163]\\
LAML & ranger & \textbf{0.709} & [0.651, 0.766] & 0.596 & [0.534, 0.657] & ranger & \textbf{0.182} & [0.165, 0.199] & 0.231 & [0.200, 0.263]\\
LGG & glmboost & \textbf{0.749} & [0.719, 0.779] & 0.652 & [0.618, 0.685] & Lasso & \textbf{0.145} & [0.132, 0.157] & 0.168 & [0.154, 0.181]\\
LIHC & grridge & \textbf{0.602} & [0.560, 0.645] & 0.586 & [0.542, 0.630] & ranger & \textbf{0.146} & [0.135, 0.157] & 0.169 & [0.158, 0.180]\\
LUAD & prioritylasso & \textbf{0.665} & [0.640, 0.690] & 0.663 & [0.631, 0.695] & CoxBoost favoring & \textbf{0.172} & [0.160, 0.183] & 0.172 & [0.161, 0.183]\\
LUSC & prioritylasso favoring & \textbf{0.537} & [0.502, 0.572] & 0.531 & [0.502, 0.561] & grridge & \textbf{0.210} & [0.203, 0.217] & 0.216 & [0.205, 0.227]\\
OV & prioritylasso & \textbf{0.600} & [0.582, 0.618] & 0.598 & [0.580, 0.617] & ipflasso & \textbf{0.169} & [0.163, 0.174] & 0.173 & [0.167, 0.179]\\
PAAD & prioritylasso favoring & 0.686 & [0.658, 0.714] & 0.683 & [0.655, 0.712] & Clinical only & 0.190 & [0.178, 0.202] & - & -\\
SARC & blockForest & \textbf{0.685} & [0.651, 0.720] & 0.673 & [0.637, 0.709] & glmboost & \textbf{0.179} & [0.167, 0.190] & 0.202 & [0.188, 0.217]\\
SKCM & blockForest & 0.597 & [0.556, 0.639] & 0.581 & [0.540, 0.623] & Clinical only & 0.191 & [0.185, 0.198] & - & -\\
STAD & Clinical only & 0.598 & [0.555, 0.641] & - & - & Clinical only & 0.192 & [0.182, 0.202] & - & -\\
UCEC & Clinical only & 0.686 & [0.581, 0.791] & - & - & ipflasso & 0.091 & [0.079, 0.102] & 0.092 & [0.080, 0.105]\\
\bottomrule
    \end{tabular}}{}
\end{table*}

\begin{table}[!h]
\vspace{1cm}
    \caption{Comparing naive learners and structured learners. The performance of structured learners, i.e. learners using the group structure, and naive learners are compared for every dataset. The cindex and ibrier columns show the mean performance values for the corresponding dataset and learner types. Bold values indicate better values for the given dataset.\label{tab:struc}}
    \centering
    {\begin{tabular}{@{}lllll@{}}
    \toprule
                & cindex     &       & ibrier     &     \\
        data  & structured & naive & structured & naive\\
        \midrule
        BLCA & \textbf{0.618} & 0.580 & \textbf{0.198} & 0.203\\
        BRCA & \textbf{0.598} & 0.467 & \textbf{0.152} & 0.180\\
        COAD & \textbf{0.518} & 0.452 & \textbf{0.104} & 0.114\\
        ESCA & \textbf{0.506} & 0.405 & 0.235 & \textbf{0.233}\\
        HNSC & \textbf{0.562} & 0.549 & \textbf{0.210} & 0.214\\
        KIRC & \textbf{0.721} & 0.671 & \textbf{0.154} & 0.159\\
        KIRP & \textbf{0.560} & 0.527 & \textbf{0.132} & 0.133\\
        LAML & \textbf{0.634} & 0.548 & \textbf{0.207} & 0.211\\
        LGG & 0.695 & \textbf{0.721} & 0.169 & \textbf{0.158}\\
        LIHC & \textbf{0.566} & 0.530 & 0.171 & \textbf{0.164}\\
        LUAD & \textbf{0.636} & 0.519 & \textbf{0.181} & 0.199\\
        LUSC & \textbf{0.501} & 0.406 & \textbf{0.220} & 0.231\\
        OV & \textbf{0.575} & 0.393 & \textbf{0.172} & 0.189\\
        PAAD & \textbf{0.663} & 0.581 & \textbf{0.196} & 0.205\\
        SARC & \textbf{0.667} & 0.624 & 0.197 & \textbf{0.180}\\
        SKCM & \textbf{0.580} & 0.478 & \textbf{0.200} & 0.221\\
        STAD & \textbf{0.556} & 0.504 & \textbf{0.199} & 0.211\\
        UCEC & \textbf{0.646} & 0.526 & \textbf{0.103} & 0.115\\
        \bottomrule
    \end{tabular}}{}
\end{table}

\section{Discussion}
In general, one should be very careful when interpreting the results of our benchmark experiment and drawing conclusions. 
Most importantly, the findings highly depend on the considered prediction performance measure,
as the method ranking changes drastically between the two measures. For example, \textit{CoxBoost} performs poorly based on the cindex but performs third best regarding the ibrier. These findings indicate that the performance of a method may change dramatically if a different performance measure is used for its assessment. Moreover, according to ibrier, two methods perform better than the Cox model (though only slightly), and six methods perform worse than the Kaplan-Meier estimate. 

Another important aspect of the performance assessment is, as shown in Figure \ref{fig:perf} and Table \ref{tab:group}, the variability across datasets. The superiority of one method over the other not only strongly depends on the considered performance measure but also, most importantly, on the considered datasets. This stresses the importance of {\it large-scale} benchmarks, like this one, which use many datasets. Since the variability between datasets is huge, we need many datasets---a fact well-known by statisticians performing sample size calculations, which however tends to be ignored when designing benchmark experiments using real datasets \citep{BoulesteixHable2015b}. If we had conducted our study with, say, 3, 5 or 10 datasets (as usual in the literature), we would have obtained different---more unstable---results.

Regarding prediction performance, \textit{blockForest} outperforms the other methods on average over all datasets. Moreover, it is the only method which outperforms the simple Cox model on average regarding both measures. The other methods using the molecular data do not perform better than the simple Cox model. The better prediction performance of \textit{blockForest}, however, comes at the price of long computation times.  Apart from \textit{SGL}, \textit{blockForest} is the slowest method. The fastest learners, standard \textit{Lasso} and \textit{ranger}, are about 10 times faster and \textit{blockForest} is still 2 times slower than the next slowest learner \textit{prioritylasso favoring}. Moreover, like the other random forest variants, it does not yield easily interpretable models, even though the strengths of the variables can be assessed via the variable importance measure(s) output as a by-product of the random forest algorithms. Thus, taking the other assessment dimensions into account, e.g. CoxBoost favoring clinical features is very competitive. It is quite fast, leads to reasonably sparse models at group and feature level and yields performances only slightly worse than (cindex) or equal to (ibrier) \textit{blockForest}.

From a practical perspective, even simpler modeling approaches, such as a simple Cox model using clinical variables only, might be preferable. This model is easily interpretable, needs only a fraction of the computation time and, with a mean cindex of 0.618 and a mean ibrier of 0.175, performs only slightly worse than \textit{blockForest} (0.620 and 0.172) and comparable to or even better than all other methods. Note, however, that \textit{blockForest} also offers the possibility of favoring clinical covariates using the argument \textit{always.select.block} of the \textit{blockfor} function. \cite{Hornung2018}, show that this can improve the prediction performance of block forest considerably. However, since this option was not yet available at the time of conduction of the analyses performed for the current paper, we were not able to consider this block forest variant here.

In general, conclusions about the superiority of one method over the other with respect to the prediction performance must be drawn with caution, as the differences in performance can be very small and the confidence intervals often show a remarkable overlap. Exemplary t-tests comparing \textit{blockForest} with \textit{CoxBoost favoring} and the \textit{Clinical only} model showed no significant differences in performance.

More generally, it should be noted that the choice of a method should result from the simultaneous consideration of various aspects beyond performance. If (i) performance is the main criterion, (ii) the model is intended solely as a prediction tool, and  implemented, say, as a \textit{shiny} application \citep{shiny}, and (iii) sparsity and interpretability are not considered important, \textit{blockForest} is certainly a very good choice. Other methods may prove attractive in different situations. Finally, let us note that one of the methods that did not perform very well in the present study in terms of performance, priority-Lasso, may perform better in practice when accurate prior knowledge on the groups of variables is available, and allows the user to favor some of the groups---a dimension that could not be taken into account in our comparison study.

A potential limitation of our study is that the datasets were already used by \cite{Hornung2018}  in their comparison study. Since they selected the most promising \textit{blockForest} based on this comparison study, our results may be slightly optimistic regarding the performance of \textit{blockForest}---a bias mechanism that has been previously described \citep{jelizarow2010}. More precisely, \cite{Hornung2018} initially considered five different variants of random forest taking the block structure into account and identified the best-performing variant using a collection of 20 datasets including the 18 considered in our study. They named this best-performing variant \lq\lq block forest''.  It is in theory possible that part of the superiority of the selected \lq\lq block forest'' variant on the specific 20 datasets is due to chance. In this case, our study that uses 18 of these datasets again would (slightly) favor block forest.
However, \cite{Hornung2018} used two additional datasets, did not use the same sets of clinical variables, and---to reduce the computational burden---, in the cases of groups with more than 2,500 variables, they used only a subset of 2,500 variables. To sum up, the benchmark by \cite{Hornung2018} and ours are largely different even if partly based on data from the same cancer studies. It is thus unlikely that our study is noticeably biased, although such a bias is possible in principle.

Regarding the advantage of favoring the clinical variables, it is important to note that it strongly depends on the level of predictive information contained in these variables. If clinical variables contain less information than for the datasets used in our analysis, favoring of these covariates might be less useful than they were found to be in this study, or even detrimental. While we strongly recommend considering favoring the clinical variables, this should not necessarily be performed by default. 

Extending the benchmark to further methods (e.g., methods that do not rely on the proportional hazards assumption, which are only represented by random forest in our study) and further data pre-processing approaches as well as further datasets are desirable.  In the same vein, it may be interesting to consider alternative procedures to handle model failures in the outer-resampling process, which may lead to different results. There is to date no widely used standardized approach to deal with NAs in this context. This issue certainly deserves further dedicated research. This benchmark experiment is designed such that such extensions are easy to implement. Using the provided code, further methods can be compared to the ones included in this study.

\section*{Funding}

This work was supported by the German Federal Ministry of Education and Research (BMBF) [01IS18036A]; and by the German Research Foundation [BO3139/4-3 to A.-L.B., HO6422/1-2 to R.H.]. The authors of this work take full responsibilities for its content. \vspace*{-12pt}

\section*{Acknowledgements}
We thank Mark van de Wiel, Benjamin Hofner, Marvin Wright, and Harald Binder for their evaluations of our code and their helpful advice regarding the specific method configurations. We also thank Alethea Charlton for proof-reading the manuscript.


%
%

\begin{thebibliography}{99}

\bibitem[Hasin {\it et~al}., 2017]{Hasin2017}
Hasin Y, Seldin M, Lusis A. Multi-omics approaches to disease. {\it Genome Biol.} 2017;\textbf{18}:83--98.

\bibitem[Boulesteix and Sauerbrei, 2011]{Boulesteix2011}
Boulesteix A-L, Sauerbrei W. Added predictive value of high-throughput molecular data to clinical data and its validation. \textit{Brief Bioinform} 2011;\textbf{12}:215--229.

\bibitem[Klau {\it et~al}., 2018]{Klau2018}
Klau S, Jurinovic V, Hornung R, \textit{et~al}. Priority-Lasso: A simple
hierarchical approach to the prediction of clinical outcome using
multi-omics data. \textit{BMC Bioinformatics} 2018;\textbf{19}:1.

\bibitem[B{\o}velstad {\it et~al}., 2009]{Bovelstad2009} 
B{\o}velstad HM, Nyg{\aa}rd S, Borgan {\O}. Survival prediction from
clinico-genomic models---a comparative study. \textit{BMC Bioinformatics} 2009;\textbf{10}:413.

\bibitem[Zhao {\it et~al}., 2014]{Zhao2014}
Zhao Q, Shi X, Xie Y, \textit{et~al}. Combining multidimensional genomic
measurements for predicting cancer prognosis: Observations from TCGA.
\textit{Brief Bioinform} 2014;\textbf{16}:291--303.

\bibitem[Lang {\it et~al}., 2015]{LangKotthaus2015} 
Lang M, Kotthaus H, Marwedel P, \textit{et al}. Automatic model selection for high-dimensional survival analysis. \textit{J Stat Comput Simul} 2015;\textbf{85}:62--76.

\bibitem[De Bin \textit{et~al}., 2019]{Bin2019}
De Bin R, Boulesteix A-L, Benner A, \textit{et~al}. Combining clinical and molecular data in regression prediction models: insights from a
simulation study. \textit{Brief Bioinform} 2019.

\bibitem[Boulesteix \textit{et~al}., 2019]{Boulesteix2019} 
Boulesteix A-L, Janitza S, Hornung R, \textit{et~al}. Making complex prediction rules applicable for readers: Current practice in random forest literature and recommendations. \textit{Biometrical J} 2019;\textbf{61}:1314--1328

\bibitem[De Bin \textit{et~al}., 2014]{BinHerold2014}
De Bin R, Herold T, Boulesteix A-L. Added predictive value of omics data: Specific issues related to validation illustrated by two case studies. \textit{BMC Med Res Method} 2014;\textbf{14}:117.

\bibitem[De Bin \textit{et~al}., 2014]{BinSauerbrei2014}
De Bin R, Sauerbrei W, Boulesteix A-L. Investigating the prediction ability of survival models based on both clinical and omics data: Two case studies. \textit{Stat Med} 2014;\textbf{33}:5310--5329.

\bibitem[Binder and Schumacher, 2008]{Binder2008}
Binder H, Schumacher M. Allowing for mandatory covariates in boosting estimation of sparse high-dimensional survival models. \textit{BMC Bioinformatics} 2008;\textbf{9}:14.

\bibitem[Tibshirani, 1996]{Tibshirani1996} Tibshirani R. Regression shrinkage and selection via the lasso. \textit{J R Stat Soc B} 1996;\textbf{58}:267--288.

\bibitem[Tibshirani, 1997]{Tibshirani1997} Tibshirani R. The lasso method for variable selection in the Cox model. \textit{Stat Med} 1997;\textbf{16}:385--395.

\bibitem[Schulze, 2017]{Schulze2017}
Schulze G. Clinical outcome prediction based on multi-omics data: Extension of IPF-LASSO. 2017.

\bibitem[Boulesteix \textit{et~al}., 2017]{Boulesteix2017} 
Boulesteix A-L, De Bin R, Jiang X, \textit{et~al}. IPF-LASSO: Integrative L1-penalized regression with penalty factors for prediction based on multi-omics data. \textit{Comput Math Methods Med} 2017. 

\bibitem[Simon \textit{et~al}., 2013]{Simon2013} 
Simon N, Friedman JH, Hastie T. A sparse-group lasso. \textit{J Comput Graph Stat} 2013;\textbf{22}:231--245.

\bibitem[Yuan and Lin, 2006]{Yuan2006} 
Yuan M, Lin Y. Model selection and estimation in regression with grouped variables. \textit{J R Stat Soc B} 2006;\textbf{68}:49--67.

\bibitem[Wiel \textit{et~al}., 2015]{Wiel2015} 
Wiel MA van de, Lien TG, Verlaat W, \textit{et~al}. Better prediction by use of co-data: Adaptive group-regularized ridge regression. \textit{Stat Med};\textbf{35}:368--381.

\bibitem[Friedman, 2001]{Friedman2001}
Friedman JH. Greedy function approximation: A gradient boosting machine. \textit{Ann Stat} 2001;\textbf{29}:1189--1232.

\bibitem[Hothorn and B{\"u}hlmann, 2006]{Hothorn2006} 
Hothorn T, B{\"u}hlmann P. Model-based boosting in high dimensions. \textit{Bioinformatics} 2006;\textbf{22}:2828--2829.

\bibitem[B{\"u}hlmann and Hothorn, 2007]{Buehlmann2007}
B{\"u}hlmann P, Hothorn T. Boosting algorithms: Regularization, prediction and model fitting. \textit{Stat Sci} 2007;\textbf{22}:477--505.

\bibitem[Tutz and Binder, 2006]{Tutz2006} 
Tutz G, Binder H. Generalized additive modeling with implicit variable selection by likelihood-based boosting. \textit{Biometrics} 2006;\textbf{62}:961--971.

\bibitem[Breiman, 2001]{Breiman2001} Breiman L. Random forests. \textit{Mach Learn} 2001\textbf{45}:5--32.

\bibitem[Ishwaran \textit{et~al}., 2008]{Ishwaran2008} 
Ishwaran H, Kogalur UB, Blackstone EH, \textit{et al}. Random survival forests. \textit{Ann Appl Stat} 2008;\textbf{2}:841--860.

\bibitem[Hornung and Wright, 2019]{Hornung2018} 
Hornung R, Wright MN. Block Forests: Random forests for blocks of clinical and omics covariate data. \textit{BMC Bioinformatics} 2019;\textbf{20}:358.

\bibitem[Boulesteix \textit{et~al}., 2013]{BoulesteixLauer2013} 
Boulesteix A-L, Lauer S, Eugster MJA. A plea for neutral comparison studies in computational sciences. \textit{PLoS ONE} 2013;\textbf{8}:e61562.

\bibitem[Boulesteix \textit{et~al}., 2017]{BoulesteixWilson2017} 
Boulesteix A-L, Wilson R, Hapfelmeier A. Towards evidence-based computational statistics: Lessons from clinical research on the role and design of real-data benchmark studies. \textit{BMC Med Res Method} 2017;\textbf{17}:138.

\bibitem[Couronn{\'{e}} \textit{et~al}., 2018]{Couronne2018} 
Couronn{\'{e}} R, Probst P, Boulesteix A-L. Random forest versus logistic regression: A large-scale benchmark experiment. \textit{BMC Bioinformatics} 2018;\textbf{19}:270.

\bibitem[Bischl \textit{et~al}., 2016]{mlr} Bischl B, Lang M, Kotthoff L, et al. mlr: Machine learning in {R}. \textit{J Mach Learn Res} 2016;\textbf{17}:1--5.

\bibitem[R Core Team, 2014]{RCore2014} R Core Team. {R}: A language and environment for statistical computing. \textit{R Foundation for Statistical Computing, Vienna, Austria} 2018. \url{https://www.R-project.org/} (17 October 2019, date last accessed).

\bibitem[Vanschoren \textit{et~al}., 2013]{OpenML2013} 
Vanschoren J, Rijn JN van, Bischl B, \textit{et~al}. OpenML: Networked science in machine learning. \textit{SIGKDD Explor} 2013;\textbf{15}:49--60.

\bibitem[Casalicchio \textit{et~al}., 2017]{OpenMLR2017} 
Casalicchio G, Bossek J, Lang M, \textit{et~al}. OpenML: An R package to connect to the machine learning platform OpenML. \textit{Comp Stat} 2017;\textbf{32}:1--15.

\bibitem[Microsoft Corporation, 2018]{checkpoint} 
Microsoft Corporation. Checkpoint: Install packages from snapshots on the checkpoint server for reproducibility. 2018.

\bibitem[Lang \textit{et~al}., 2017]{batchtools} 
Lang M, Bischl B, Surmann D. Batchtools: Tools for {R} to work on batch systems. \textit{J Open Source Softw} 2017;\textbf{2}:135.

\bibitem[Uno \textit{et~al}., 2011]{Uno2011} 
Uno H, Cai T, Pencina MJ, \textit{et~al}. On the C-statistics for evaluating overall adequacy of risk prediction procedures with censored survival data. \textit{Stat Med} 2011;\textit{30}:1105--1117.

\bibitem[Friedman \textit{et~al}., 2010]{glmnet} 
Friedman JH, Hastie T, Tibshirani R. Regularization paths for generalized linear models via coordinate descent. \textit{J Stat Softw} 2010;\textbf{33}:1--22.

\bibitem[Simon \textit{et~al}., 2011]{Simon2011} 
Simon N, Friedman JH, Hastie T, \textit{et~al}. Regularization paths for Cox's proportional hazards model via coordinate descent. \textit{J Stat Softw} 2011;\textbf{39}:1--13.

\bibitem[Simon \textit{et~al}., 2018]{SGL} 
Simon N, Friedman J, Hastie T, \textit{et~al}. SGL: Fit a glm (or Cox model) with a combination of lasso and group lasso regularization 2018. \url{https://CRAN.R-project.org/package=SGL} (17 October 2019, date last accessed).

\bibitem[Boulesteix and Fuchs, 2015]{ipflasso} 
Boulesteix A-L, Fuchs M. Ipflasso: Integrative lasso with penalty factors 2015. \url{https://CRAN.R-project.org/package=ipflasso} (17 October 2019, date last accessed).

\bibitem[Klau and Hornung, 2017]{prioritylasso} 
Klau S, Hornung R. Prioritylasso: Analyzing multiple omics data with an offset approach 2017. \url{https://CRAN.R-project.org/package=prioritylasso}(17 October 2019, date last accessed).

\bibitem[Wiel and Novianti, 2018]{ggridge} 
Wiel MA van de, Novianti PW. GRridge: Better prediction by use of co-data: Adaptive group-regularized ridge regression 2018. \url{https://bioconductor.org/packages/release/bioc/html/GRridge.html} (17 October 2019, date last accessed).

\bibitem[Hothorn \textit{et~al}., 2018]{mboost} 
Hothorn T, B{\"u}hlmann P, Kneib T, \textit{et~al}. mboost: Model-based boosting 2018. \url{https://CRAN.R-project.org/package=mboost} (17 October 2019, date last accessed).

\bibitem[Binder, 2013]{CoxBoost} 
Binder H. CoxBoost: Cox models by likelihood based boosting for a single survival endpoint or competing risks 2013. \url{https://CRAN.R-project.org/package=CoxBoost} (17 October 2019, date last accessed).

\bibitem[Ishwaran and Kogalur, 2018]{RFSRC} 
Ishwaran H, Kogalur UB. Random forests for survival, regression, and classification (rf-src) 2018. \url{https://CRAN.R-project.org/package=randomForestSCR} (17 October 2019, date last accessed).

\bibitem[Wright and Ziegler, 2017]{ranger}
Wright MN, Ziegler A. ranger: A fast implementation of random forests for high dimensional data in C++ and {R}. \textit{J Stat Softw} 2017;\textbf{77}:1--17.

\bibitem[Hornung and Wright, 2019]{blockForest} 
Hornung R, Wright MN. BlockForest: Block forests: Random forests for blocks of clinical and omics covariate data 2019. \url{https://CRAN.R-project.org/package=blockForst} (17 October 2019, date last accessed).

\bibitem[Therneau, 2015]{survival} 
Therneau TM. A package for survival analysis in {S} 2015. \url{https://CRAN.R-project.org/package=survival} (17 October 2019, date last accessed).

\bibitem[Bischl \textit{et~al}., 2013]{BischlSchiffner2013} 
Bischl B, Schiffner J, Weihs C. Benchmarking local classification methods. \textit{Comp Stat} 2013;\textbf{28}:2599--2619.

\bibitem[Probst \textit{et~al}., 2019]{Probst2019}
Probst P, Wright M, Boulesteix A-L. Hyperparameters and tuning strategies for random forest. \textit{WIRES Data Min Knowl} 2019;\textbf{9}:e1301.

\bibitem[De Bin, 2016]{Bin2016} 
De Bin R. Boosting in Cox regression: A comparison between the likelihood-based and the model-based approaches with focus on the R-packages CoxBoost and mboost. \textit{Comp Stat} 2016;\textbf{31}:513--531.

\bibitem[Boulesteix \textit{et~al}., 2015]{BoulesteixHable2015b} 
Boulesteix A-L, Hable R, Lauer S, \textit{et~al}. A statistical framework for hypothesis testing in real data comparison studies. \textit{Am Stat} 2015;\textbf{69}:201--212.

\bibitem[Chang \textit{et~al}., 2018]{shiny} 
Chang W, Cheng J, Allaire JJ, \textit{et~al}. Shiny: Web application framework for {R} 2018. \url{https://CRAN.R-project.org/package=shiny} (12 January 2020, date last accessed).

\bibitem[Jelizarow \textit{et~al}., 2010]{jelizarow2010} 
Jelizarow M, Guillemot V, Tenenhaus A, \textit{et~al}. Over-optimism in bioinformatics: An illustration. \textit{Bioinformatics} 2010;\textbf{26}:1990--1998.


\end{thebibliography}

\renewcommand{\bibnumfmt}[1]{#1. }

\end{document}


\section*{Supplement: Method details}
Let there be a feature matrix \textbf{X} and $G$ feature groups, then $\bm{x}^{(g)}_{\cdot 1}, ... , \bm{x}^{(g)}_{\cdot p_g}$ denote the features of group $g$, $g = 1,...,G$, with $p_g$ the number of features within group $g$, and $\bm{x}_i$ the feature vector of observation $i$. Furthermore $\beta^{(g)}_j$ indicates the corresponding coefficient of feature $x^{(g)}_{\cdot j}$. 

For survival outcomes the penalized-regression and the boosting methods are based on the concept of minimizing the negative partial log-likelihood 
\begin{equation}\label{eq:loglike}
    pl(\bm\beta) = -\sum^n_{i=1}\delta_i \left[\bm{x}_i^T\bm\beta - log\left(\sum_{l \in R_i} exp(\bm{x}_l^T\bm\beta)\right)\right]
\end{equation}
rather than minimizing the sum of squared differences,  
with $R_i$ the set of indices of individuals at risk at time $t_i$ and $\delta_i \in \{0, 1\}$ the censoring indicator \citep{Tibshirani1997, Binder2008}. 
If an offset term $\eta_i$ is to be included in the estimation process, (\ref{eq:loglike}) becomes
\begin{equation}\label{eq:loglike2}
    pl(\bm\beta\vert\eta) = -\sum^n_{i=1}\delta_i\left[\eta_i + \bm{x}_i^T\bm\beta - log\left(\sum_{l \in R_i} exp(\bm{x}_l^T\bm\beta)\right)\right]
\end{equation}\\

\subsection*{Penalized-regression-based methods}
For standard \textit{Lasso} \citep{Tibshirani1996, Tibshirani1997}, which does not regard the group structure, the estimate in the survival setting is given by    
\begin{equation}
    \hat{\bm\beta} = \argmin_{\bm\beta} pl(\bm\beta) + \lambda\sum^{p}_{j=1}|\beta_j|.
\end{equation} 
Large coefficient values are penalized and sparse final models result since a substantial number of coefficients are set to zero, if the penalty parameter $\lambda$ is large enough, which is a hyper-parameter to be tuned. \\

\textit{Two-step} (TS) \textit{IFP-Lasso} \citep{Schulze2017, Boulesteix2017} is an extension of the standard Lasso specifically designed to incorporate multi-omics group structure and based on the estimate
\begin{equation}
    \hat{\bm\beta} = \argmin_{\bm{\beta}}  pl(\bm\beta) + \sum^G_{g=1}\lambda_g \Vert\bm{\beta}^{(g)}\Vert_1,
\end{equation}
with $\Vert.\Vert_1$ being the $L_1$-Norm, $\lambda_g > 0$ the penalty of group $g$ and $\bm{\beta}^{(g)} = (\beta^{(g)}_1,..., \beta^{(g)}_{p_g})^T$.
It uses different penalties for the specified groups by determining a vector of inverted coefficient means for every group which are used as penalty factors. These penalty factors are determined in the first step by fitting a preliminary model.\\
\\
\textit{Priority-Lasso} \citep{Klau2018}  successively fits regression models with the estimate
\begin{equation}
    \hat{\bm\beta}^{(\pi_g)} = \argmin_{\bm\beta^{(\pi_g)}} pl(\bm\beta^{(\pi_g)}\vert \hat{\eta}^{(\pi_g-1)}) + \lambda^{(\pi_g)}\sum^{p_{\pi_g}}_{j =1}|\beta^{(\pi_g)}_j|
\end{equation}
\noindent with $\bm\pi = (\pi_1,...,\pi_G)$ being a permutation of $(1,...,G)$ that indicates the priority order and $\hat{\eta}^{(\pi_{g-1})}$ 
the resulting linear predictor of group $\pi_{g-1}$. More precisely, the single estimates $\hat{\bm\beta}^{(\pi_g)}$ are obtained by sequentially fitting Lasso models using the features in the order of their group's priority. The linear predictor $\hat{\eta}_i^{(\pi_{g-1})}$ of the previous group $\pi_{g-1}$ is used as offset for the model fit to the next group.\\
\\
\textit{Sparse Group Lasso} (SGL) \citep{Simon2013} is another extension of the Lasso capable of including group information. The estimate is given by
\begin{equation}
    \hat{\bm\beta} = \argmin_{\bm\beta} \frac{1}{n} pl(\bm\beta) + (1-\alpha)\lambda \sum^G_{g=1}\sqrt{p_g}\vert\vert\bm\beta^{(g)}\vert\vert_2 + \alpha\lambda\vert\vert\bm\beta\vert\vert_1
\end{equation}
\noindent with $\alpha$ steering the contributions of the group-Lasso \citep{Yuan2006} and the standard Lasso penalties. This simultaneously leads to sparsity on feature as well as on group level.\\
\\
\textit{Adaptive group-regularized ridge regression} (GRridge) \citep{Wiel2015} is designed to use group specific co-data, e.g. p-values known from previous studies, but multi-omics group structure may also be regarded as co-data. The estimate is given by 
\begin{equation}
    \hat{\bm\beta} = \argmin_{\bm\beta} pl(\bm\beta) + \sum^{G}_{g=1} \lambda_g \sum^{p_g}_{j = 1} \bm\beta_j^2 
\end{equation}
\noindent with $\lambda_g = \lambda^{'}_g\lambda$ and $\lambda$ a global penalty, specified by cross-validation, and $\lambda^{'}_g$ factors which are estimated via empirical Bayes estimation. Feature selection is achieved post-hoc by exploiting the heavy-tailed distribution of the estimated coefficients, which clearly separates coefficients close to zero from those further away \citep{Wiel2015}. \\

\subsection*{Boosting-based methods}
Boosting is a general technique introduced in the context of classification in the machine learning community, which has then been transferred to statistical contexts as a gradient descent in function space \citep{Friedman2001}. Statistical boosting can thus be seen as a form of iterative function estimation by fitting a series of weak models, so called base learners. In general, one is interested in a function $f$ that minimizes the expected loss when used to model the data. Given a target variable $Y$ and a matrix of independent variables $\mathcal{X}$, this can be expressed as 
\begin{equation}
    f^*(.) = \argmin_{f(.)} \mathbf{E}[\rho(Y, f(\mathcal{X}))],
\end{equation}
with $\rho$ being a loss function \citep{Buehlmann2007}. The target function $f^*(.)$ is updated iteratively, with the number of boosting steps $m_{stop}$, i.e. iterations, being the main tuning parameter. This results in a model 
\begin{equation}
    \hat{f}^*(\textbf{X}) = \sum^{m_{stop}}_{m=0} \hat{f}^{[m]}(\textbf{X}) = \hat{f}^{[0]}(\textbf{X}) + \nu\sum^{m_{stop}}_{m=1} \hat{\mathcal{G}}^{[m]}(\bm{u}^{[m]}, \textbf{X}),
\end{equation} 
with $\bm{u}^{[m]} = -\frac{\delta}{\delta f}\rho(\bm{y}, f)|_{f=\hat{f}^{[m-1]}(\textbf{X})}$ being the negative gradient, $\mathcal{G}$ the base learner and $\nu$ the learning rate \citep{Hastie2009}.\\
\\ 
\textit{Model-based boosting} uses simple linear models as base learners and updates only the loss minimizing base learner per iteration, leading to an estimate where each component depends only on one feature, i.e.
\begin{equation}
    \hat{f}^*(\textbf{X}) = \sum^{m_{stop}}_{m=0} \hat{f}^{[m]}(\textbf{X}) = \hat{f}^{[0]}(\textbf{X}) + \nu\sum^{m_{stop}}_{m=1} \hat{\bm\beta}^{[m]}_j \bm{x}_{.j},
\end{equation}
with $\hat{\bm\beta}^{[m]}_j = (\bm{x}^T_{\cdot j}\bm{x}_{\cdot j})^{-1}\bm{x}^T_{\cdot j}\hat{\bm{u}}^{[m]}$. In the survival setting the negative gradient for observation $i$ becomes \begin{equation}
    \hat{u}_i^{[m]} = \delta_i - \sum_{l \in R_i} \delta_l \frac{exp(\bm{x_l}^T\hat{\bm\beta}^{[m-1]})}{\sum_{k \in R_l} exp(\bm{x}_k^T\hat{\bm{\beta}}^{[m-1]})}.
\end{equation}

\noindent \textit{Likelihood-based boosting} \citep{Tutz2006}, in contrast, uses a penalized version of the partial log-likelihood as loss and the shrinkage is directly applied in the coefficient estimation step via a penalty parameter, leading to 
\begin{equation}\label{eq:likelihood}
    pl_{LB}(\bm\beta^{[m]}) = pl(\bm\beta^{[m]}|\hat{\bm{\eta}}^{[m-1]}) - 0.5 \lambda \bm\beta^T\textbf{P}\bm\beta,
\end{equation}
where $\textbf{P}$ is a $p  \times p$ matrix \citep{Bin2016}. As it is still an iterative procedure, the updates of previous iterations have to be included as an offset $\hat{\bm\eta}^{[m-1]} = \bm{x}^T\hat{\bm\beta}^{[m-1]}$ in expression \eqref{eq:likelihood}, to make use of the information gained.\\

\subsection*{Random-forest-based methods}
The following algorithm is based on \citet[p. 588]{Hastie2009}.
\noindent To compute a \textit{random forest}\\
\\
1. For $b = 1$ to $B$:
\\\\
\indent 1.1 Draw a bootstrap sample of size $n$ from the training data\\\\
\indent 1.2  Fit a single tree $T_b$ to the bootstrap sample by recursively repeat- \\
\indent \indent ing the following steps for each node, until the predefined\\
\indent \indent minimum $nodesize$\\\\
\indent \indent a) Draw $mtry$ numbers of features randomly \\
\indent \indent b) Choose the best feature and split-point combination\\
\indent \indent c) Split the node into two daughter nodes\\\\
2. Output the tree ensemble\\
\\
The extension to survival time data by \citet{Ishwaran2008} is also called \textit{random survival forest}. The split criterion is based on maximizing the difference in survival. The terminal $nodesize$ is defined by the minimum number of deaths $d_0 > 0$ that should be included in the final nodes $\mathcal{T}$ of the trees constituting the forest. Let $t_1^{(\mathcal{T})}$ < $t_2^{(\mathcal{T})}$ <  ... < $t_{N_{\mathcal{T}}}^{(\mathcal{T})}$be the $N_{\mathcal{T}}$ event times in a final node $\mathcal{T}$, for each $\mathcal{T}$ the cumulative hazard function $\Lambda^{(\mathcal{T})}(t^*) = \int_0^{t^*}\lambda(u)du$, $\lambda(u)$ the hazard rate at time point $u$, is computed via the Nelson-Aalen estimator
\begin{equation}
    \hat\Lambda^{(\mathcal{T})}(t^*) = \sum_{t_l^{(\mathcal{T})} \leq t^*} \frac{d_{l}^{(\mathcal{T})}}{n_l^{(\mathcal{T})}},
\end{equation} with $d^{(\mathcal{T})}_l$ the number of deaths and $n_l^{(\mathcal{T})}$ = $\vert R^{(\mathcal{T})}_l\vert$ the number of individuals at risk at time $t_l^{(\mathcal{T})}$. For prediction, the estimates are averaged across the trees to obtain the ensemble cumulative hazard function.

\textit{Block forest} \citep{Hornung2018} modifies the split point selection of random (survival) forest to incorporate the block structure of multi-omics data. The block choice is randomized and the split criterion values are weighted blockwise in the split-point selection procedure.

\renewcommand{\bibnumfmt}[1]{#1. }